\pdfoutput=1
\documentclass[11pt]{article}

\usepackage{acl}

\usepackage{times}
\usepackage{latexsym}

\usepackage[T1]{fontenc}
\usepackage[utf8]{inputenc}

\usepackage{microtype}

\usepackage{url}
\usepackage{float,cuted}
\usepackage{graphicx}
\usepackage{subfigure}
\usepackage{amssymb}
\usepackage{amsmath,bm}
\usepackage{paralist,tabularx,multirow}
\usepackage{caption}
\usepackage{booktabs}
\usepackage{makecell}
\usepackage{enumerate}
\usepackage{enumitem}
\usepackage{bbding}

\usepackage[markup=underlined]{changes}
\title{AdaTranS: Adapting with Boundary-based Shrinking for End-to-End \\Speech Translation}

\author{Xingshan Zeng, Liangyou Li, Qun Liu \\
  Huawei Noah's Ark Lab \\
  \texttt{\{zeng.xingshan,liliangyou,qun.liu\}@huawei.com} \\}

\begin{document}
\maketitle

\begin{abstract}
% End-to-end speech translation (ST) has become popular due to its advantages of lower latency and less error propagation than cascaded models. 
To alleviate the data scarcity problem in End-to-end speech translation (ST), pre-training on data for speech recognition and machine translation is considered as an important technique. However, the modality gap between speech and text prevents the ST model from efficiently inheriting knowledge from the pre-trained models. In this work, we propose AdaTranS for end-to-end ST. It adapts the speech features with a
new shrinking mechanism to mitigate the length mismatch between speech and text features by predicting word boundaries. 
% Then, we introduce two extra training objectives to help transfer knowledge from the text domain to the speech domain. 
Experiments on the MUST-C dataset demonstrate that AdaTranS achieves better performance than the other shrinking-based methods, with higher inference speed and lower memory usage. Further experiments also show that AdaTranS can be equipped with additional alignment losses to further improve performance.
% analyses show the effectiveness of our methods in bridging the modality gap and different model architectures. 
\end{abstract}
\section{Introduction}
\label{sec:intro}
End-to-end speech translation (ST), which directly translates source speech into text in another language, has achieved remarkable progress in recent years~\cite{duong-etal-2016-attentional,DBLP:journals/corr/WeissCJWC17,DBLP:conf/icassp/BerardBKP18,wang-etal-2020-curriculum,xu-etal-2021-stacked,ye-etal-2022-cross}. Compared to the conventional cascaded systems~\cite{DBLP:conf/icassp/Ney99,DBLP:conf/icassp/MathiasB06}, the end-to-end models are believed to have the advantages of low latency and less error propagation. A well-trained end-to-end model typically needs a large amount of training data. However, the available direct speech-translation corpora are very limited~\cite{di-gangi-etal-2019-must}. Given the fact that data used for automatic speech recognition (ASR) and machine translation (MT) are much richer, the paradigm of ``pre-training on ASR and MT data and then fine-tuning on ST'' becomes one of the approaches to alleviate the data scarcity problem~\cite{bansal-etal-2019-pre,xu-etal-2021-stacked}.

It has been shown that decoupling the ST encoder into acoustic and semantic encoders is beneficial to learn desired features~\cite{DBLP:journals/corr/abs-2010-14920,zeng-etal-2021-realtrans}. Initializing the two encoders by pre-trained ASR and MT encoders, respectively, can significantly boost the performance~\cite{xu-etal-2021-stacked}. However, the modality gap between speech and text might prevent the ST models from effectively inheriting the pre-trained knowledge~\cite{xu-etal-2021-stacked}. 
% As shown in Figure~\ref{fig:intro}, the inputs of the MT encoder are text embeddings, while the inputs of the semantic encoder are much longer speech features extracted by the acoustic encoder.
% \input{figs/intro}

The modality gap between speech and text can be summarized as two dimensions. First, the length gap -- the speech features are usually much longer than their corresponding texts~\cite{DBLP:conf/nips/ChorowskiBSCB15,DBLP:journals/corr/abs-2010-14920}. Second is the representation space gap. 
% The speech features from the ASR encoder and the text embeddings for the MT encoder are independently learned in different spaces during pre-training. 
Directly fine-tuning MT parameters (semantic encoder and decoder) with speech features as inputs, which learned independently, would result in sub-optimal performance. Previous work has explored and proposed several alignment objectives to address the second gap, e.g., Cross-modal Adaption~\cite{DBLP:journals/corr/abs-2010-14920}, Cross-Attentive Regularization~\cite{tang-etal-2021-improving} and Cross-modal Contrastive~\cite{ye-etal-2022-cross}. 

A shrinking mechanism is usually used to address the length gap.
Some leverage Continuous Integrate-and-Fire (CIF)~\cite{DBLP:conf/icassp/Dong020} to shrink the long speech features~\cite{dong-etal-2022-learning,DBLP:conf/interspeech/ChangL22}, but they mostly work on simultaneous ST and need extra efforts to perform better shrinking. Others mainly depend on the CTC greedy path~\cite{DBLP:journals/corr/abs-2010-14920,gaido-etal-2021-ctc}, which might introduce extra inference cost and lead to sub-optimal shrinking results. AdaTranS uses a new shrinking mechanism called boundary-based shrinking, which achieves higher performance.

Through extensive experiments on the MUST-C~\cite{di-gangi-etal-2019-must} dataset, we show that AdaTranS is superior to other shrinking-based methods with a faster inference speed or lower memory usage. Further equipped with alignment objectives, AdaTranS shows competitive performance compared to the state-of-the-art models.
% We conduct experiments on MUST-C~\cite{di-gangi-etal-2019-must} En--De, En--Fr and En--Ru datasets. The experimental results show that AdaTranS performs the best among different adaptation methods targeting on reducing modality gap between speech and text. Further experiments show that our boundary-based shrinking method are superior to other related works with a faster inference speed or lower memory usage; our modality alignment objective yields better performance.
% We also present an ablation study and further analyses to show the effectiveness and generalizability of AdaTranS.

% The contributions are summarized as follows:
% \begin{itemize}[leftmargin=*,topsep=2pt,itemsep=2pt,parsep=0pt]
% \item  We point out the drawbacks of previous shrinking methods and propose AdaTranS with a new boundary-based shrinking method for ST, which outperforms previous methods in experiments.

% \item We further propose the modality alignment to help ST models inherit knowledge from pre-trained MT models to the speech domain.

% \item AdaTranS achieves better results in the MUST-C datasets with a faster inference speed and lower memory usage than the baseline and most prior related works.
% \end{itemize}
% \input{sections/related-work.tex}
\section{Proposed Model: AdaTranS}
% In this section, we describe our proposed AdaTranS model. We first display the general formulation and architecture for ST, and then show how we shrink the speech sequences based on a boundary predictor and how to perform the modality alignment between text and speech. Finally, we summarize the final training objectives.

% \subsection{Problem Formulation}
% An ST corpus is denoted as $\mathcal{D}_{ST}=\{(\bm{x}, \bm{z}, \bm{y})\}$, containing triples of speech, transcription and translation. Here $\bm{x} = (x_1, x_2, ..., x_{T_x})$ is a sequence of speech features or waves as speech input, while $\bm{z} = (z_1, z_2, ..., z_{T_z})$ and $\bm{y} = (y_1, y_2, ..., y_{T_y})$ are the corresponding transcription in source language and translation in target language, respectively. $T_x$, $T_z$, and $T_y$ are the lengths of speech, transcription and translation, respectively, where usually $T_x \gg T_z$ and $T_x \gg T_y$. 
% We can use pre-trained ASR and MT models to initialize the components of the ST model, as a method to leverage extra data $\mathcal{D}_{ASR}=\{(\bm{x}, \bm{z})\}$ and $\mathcal{D}_{MT}=\{(\bm{z}, \bm{y})\}$.
% which are pre-trained with
% extra ASR and MT data, denoted as $\mathcal{D}_{ASR}=\{(\bm{x}, \bm{z})\}$ and $\mathcal{D}_{MT}=\{(\bm{z}, \bm{y})\}$.

\subsection{Architecture}
% An End-to-End ST model generally consists of an ST encoder and an ST decoder. 
Following previous studies~\cite{DBLP:journals/corr/abs-2010-14920,xu-etal-2021-stacked}, AdaTranS decouples the ST encoder into an acoustic encoder and a semantic encoder. 
% We define the output of the acoustic encoder as:
% \begin{equation}
% \label{eq:acoustic-enc} \small
%     \bm{H}^{A} = \ea(\bm{x})
% \end{equation}
% To make the output of the acoustic encoder consistent with the representations that work for the semantic encoder and 
To bridge the modality gap between speech and text, an adaptor is usually needed before the semantic encoder. We choose the shrinking operation~\cite{DBLP:journals/corr/abs-2010-14920,zeng-etal-2021-realtrans} as our adaptor, where the long speech sequences are shrunk to the similar lengths as the transcription based on designed mechanisms (details will be introduced in the next subsection). The shrunk representations are sent to the semantic encoder to derive 
the encoder output.
% \begin{equation}
% \label{eq:semantic-enc} \small
%     \bm{H}^{S} = \es(\shrink(\bm{H}^{A}))
% \end{equation}
Finally, the semantic output is fed into the ST decoder for computing the cross-entropy loss:
\begin{equation} \label{eq:st-loss} \small
    \mathcal{L}_{ST} = - \sum_{|\mathcal{D}_{ST}|} \, \sum_{t=1}^{T_y} \text{log} \, p(y_t | y_{<t},\bm{x})
\end{equation}

\begin{figure}[t]
\centering
\includegraphics[width=0.48\textwidth]{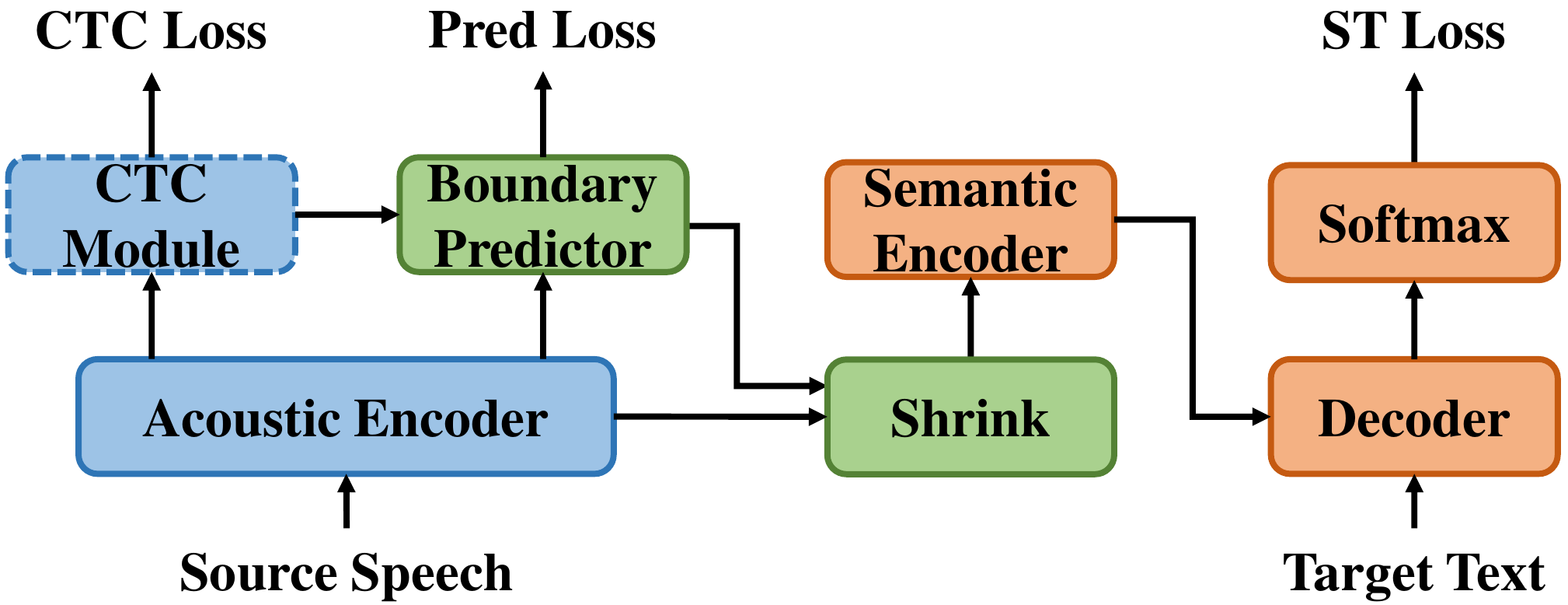}
\vskip -0.5em
\caption{
AdaTranS Architecture, where the blue modules are initialized with the ASR model and the orange modules are initialized with the MT model. The CTC module (dotted) can be removed during inference.
}
\vskip -1em
\label{fig:arch}
\end{figure}

To incorporate extra ASR and MT data, we use the pre-trained ASR encoder to initialize the ST acoustic encoder, and the pre-trained MT encoder and decoder to initialize the ST semantic encoder and decoder, respectively. Both pre-trained models are first trained with extra ASR (or MT) data and then fine-tuned with the in-domain data (the ASR part or MT part in the ST corpus). Figure~\ref{fig:arch} displays our architecture as well as the training process.

\subsection{Boundary-based Shrinking Mechanism}
\label{sec:model:shrink}
% Shrinking mechanisms have been shown to be effective in bridging modality gap between speech and text representations and so improve the performance~\cite{DBLP:journals/corr/abs-2010-14920,gaido-etal-2021-ctc,zeng-etal-2021-realtrans}. 

Previous shrinking mechanisms~\cite{DBLP:journals/corr/abs-2010-14920,zeng-etal-2021-realtrans} mostly depend on a CTC module~\cite{DBLP:conf/icml/GravesFGS06} to produce token-label probabilities for each frame in the speech representations. Then, a word boundary is recognized if the labeled tokens of two consecutive frames are different. There are two main drawbacks to such CTC-based methods. First, 
% \llyreplace{the argmax operation is usually used to extract the corresponding label of each speech frame, which might lead to a sub-optimal solution since it greedily search a CTC path. Second, they introduce extra parameters and computation cost in the CTC module during inference.}
the word boundaries are indirectly estimated and potentially affected by error propagation
from the token label predictions which are usually greedily estimated by the argmax operation on the CTC output probabilities. Second, the token labels are from a large source vocabulary resulting in extra parameters and computation cost in the CTC module during inference.

We introduce a boundary-based shrinking mechanism to address the two drawbacks. 
% \llyreplace{It uses the CTC module to guide the training of our boundary predictor (but will be discarded during inference), and then shrinks the speech representations based on the predictor output.}
A boundary predictor is used to directly predict the probability of each speech representation being a boundary, which is then used for weighted shrinking. Since the boundary labels on the speech representations are unknown during training, we introduce signals from the CTC module to guide the training of the boundary predictor. The CTC module will be discarded during inference. Below shows the details.

\paragraph{CTC module.}
We first briefly introduce the CTC module. It predicts a path $\bm{\pi} = (\pi_1, \pi_2, ..., \pi_{T_x})$, where $T_x$ is the length of hidden states after the acoustic encoder. And $\pi_t \in \mathcal{V} \cup \{\phi\}$ can be either a token in the source vocabulary $\mathcal{V}$ or the blank symbol $\phi$. By removing blank symbols and consecutively repeated labels, denoted as an operation $\mathcal{B}$, we can map the CTC path to the corresponding transcription. A CTC loss is defined as the probability of all possible paths that can be mapped to the ground-truth transcription $\bm{z}$:
\begin{equation} \label{eq:ctc-loss} \small
    \mathcal{L}_{CTC} = - \sum_{|\mathcal{D}_{ST}|} \, \sum_{\bm{\pi} \in \mathcal{B}^{-1}(\bm{z})} \text{log} \, p(\bm{\pi} | \bm{H}^{A})
\end{equation}

\begin{figure}[t]
\centering
\includegraphics[width=0.48\textwidth]{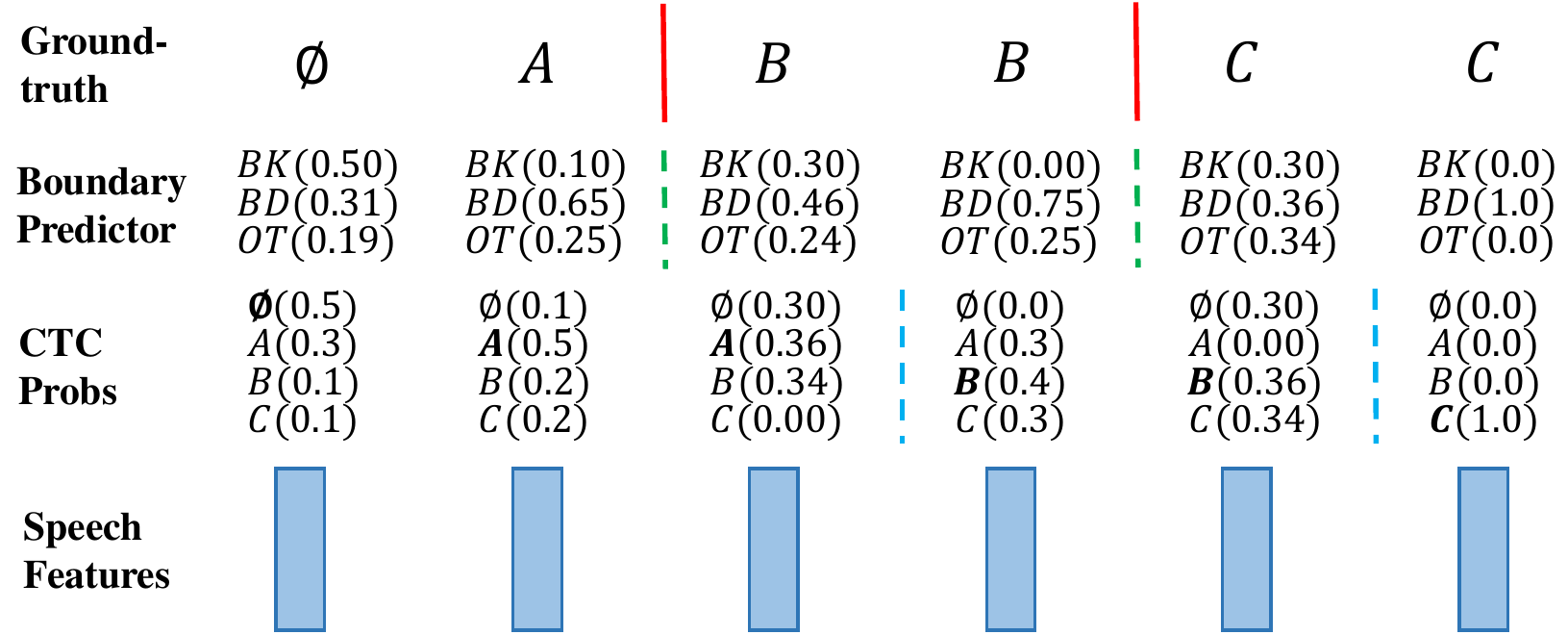}
\vskip -0.5em
\caption{
An example of CTC probabilities (we assume there are three tokens in vocabulary for simplification), together with the corresponding boundary predictor soft signals and ground labels. We will get different boundary detection results based on their label probabilities (the solid and dotted lines). CTC argmax path predicts the wrong boundaries and we can get correct boundaries if we set the threshold $\theta$ as 0.5 in boundary predictor.
}
\vskip -1em
\label{fig:ctc-path-bound}
\end{figure}

\paragraph{CTC-guided Boundary Predictor.}
% The idea behind our method comes from the observation that CTC-based shrinking methods actually do not ``care about'' what exact token each frame in speech representations is aligned to, but only need to distinguish the blank frames and boundary frames. 
% As the shrinking can be performed without knowing what exact token each frame is aligned to, 
We propose to use a boundary predictor to replace the CTC module, which has a similar architecture but with only three labels. The three labels are <BK> (blank label), <BD> (boundary label) and <OT> (others), respectively. 
However, the ground-truth labels for training the predictor are unknown. Therefore, we introduce soft training signals based on the output probabilities of the CTC module. Specifically, the ground-truth probabilities of each frame $t$ to be labeled as the three labels are defined as:
\begin{equation} \label{eq:pred-ground} \small
\begin{split}
    &p^{\prime}_t(\text{<BK>}) = p(\pi_t = \phi) \\
    &p^{\prime}_t(\text{<BD>}) = \sum\nolimits_{i \neq \phi }p(\pi_t = i)p(\pi_{t+1} \neq i) \\
    &p^{\prime}_t(\text{<OT>}) = 1 - p^{\prime}_t(\text{<BK>}) - p^{\prime}_t(\text{<BD>}) \\
\end{split}
\end{equation}
Then, the objective for the boundary predictor\footnote{Since the training of the predictor highly depends on the quality of the CTC output, the CTC module is also pre-trained.} is:
\begin{equation} \label{eq:pred-loss} \small
\setlength{\abovedisplayskip}{1pt}
\setlength{\belowdisplayskip}{1pt}
    \mathcal{L}_{Pred} = - \sum_{|\mathcal{D}_{ST}|} \sum_{t=1}^{T^\prime_x} \sum_{i \in \Delta} p^{\prime}_t(i) \; \text{log} \, p_t(i)
\end{equation}
where $\Delta = \{\text{<BK>}, \text{<BD>}, \text{<OT>}\}$.
% \llydelete{In this way, each frame's label is no longer determined by the argmax path, which potentially improves the boundary prediction. Furthermore, the}
The CTC module is only used in the training process and can be discarded during inference. Since the number of labels in the predictor is significantly smaller than the size of the source vocabulary, the time and computation costs introduced by the predictor are negligible. Figure~\ref{fig:ctc-path-bound} shows an example to elaborate the advantage of such a predictor.
% convert the CTC output probabilities to the training signals of the predictor.

\paragraph{Weighted Shrinking.}
For shrinking, we define boundary frames as those with the probabilities of the <BD> label higher than a pre-defined threshold $\theta$. The frames between two boundary frames are defined as one segment, which can be aligned to one source token. Inspired by \citet{zeng-etal-2021-realtrans}, we sum over the frames in one segment weighted by their probabilities of being blank labels to distinguish informative and non-informative frames:
\begin{equation} \label{eq:shrink}
\small
\setlength{\abovedisplayskip}{1pt}
\setlength{\belowdisplayskip}{1pt}
    \bm{h}^A_{t^{\prime}} = \sum_{t \in seg\,t^{\prime} } \bm{h}^A_t\frac{\exp(\mu (1 - p_t(\text{<BK>})))}{\sum_{s\in seg\,t^{\prime}}\exp(\mu (1 - p_s(\text{<BK>})))}
\end{equation}
where $\mu \geq 0$ denotes the temperature for the Softmax Function.

\paragraph{Forced Training.}
We introduce a forced training trick to explicitly solve the length mismatch between speech and text representations. During training, we set the threshold $\theta$ dynamically based on the length of $\bm{z}$ to make sure the shrunk representations have exactly the same lengths as their corresponding transcriptions. Specifically, we first sort the probabilities to be <BD> of all frames in descending order, and then select the $T_z$-th one as the threshold $\theta$. 

\subsection{Training Objectives}
\label{sec:model:loss}
The total loss of our AdaTranS will be:
\begin{equation} 
\label{eq:total-loss} \small
    \mathcal{L} = \mathcal{L}_{ST} + \alpha \cdot \mathcal{L}_{CTC} + \beta \cdot \mathcal{L}_{Pred}  
\end{equation}
where $\alpha$, $\beta$ are hyper parameters that control the effects of different losses.

\section{Experiments}
\subsection{Experimental Setup}
\paragraph{Datasets.}
We conduct experiments on three language pairs of MUST-C dataset~\cite{di-gangi-etal-2019-must}: English-German (En--De), English-French (En--Fr) and English-Russian (En--Ru). 
% All of them include source audios with the corresponding transcriptions in the source language and translations in the target language. 
We use the ofﬁcial data splits for train and development and tst-COMMON for test. We use LibriSpeech~\cite{DBLP:conf/icassp/PanayotovCPK15} as the extra ASR data to pre-train the acoustic model. OpenSubtitles2018\footnote{http://opus.nlpl.eu/OpenSubtitles-v2018.php} or WMT14\footnote{https://www.statmt.org/wmt14/translation-task.html} are used to pre-train the MT model. 
% For a fair comparison, we extract similar number of sentences\llynote{not clear}.
The data statistics are listed in Table~\ref{tab:stat} of Appendix~\ref{sec:appendix:data}.

\paragraph{Preprocessing.}
We use 80D log-mel filterbanks as speech input features and SentencePiece\footnote{https://github.com/google/sentencepiece}~\cite{kudo-richardson-2018-sentencepiece} to generate subword vocabularies for each language pair. Each vocabulary is learned on all the texts from ST and MT data and shared across source and target languages, with a total size of 16000. More details please refer to the Appendix~\ref{sec:appendix:imp}.

\paragraph{Model Setting.}
Conv-Transformer~\cite{DBLP:conf/interspeech/HuangHYC20} or Conformer~\cite{DBLP:conf/interspeech/GulatiQCPZYHWZW20} (results in Table~\ref{tab:main-compare-short} are achieved by AdaTranS with Conformer) is used as our acoustic encoder, both containing 12 layers. 
% More kinds of models are also tested to show the generalization capability of our method (see Section~\ref{sec:res:analysis}). 
For the semantic encoder and ST decoder, we follow the general NMT Transformer settings (i.e., both contain 6 layers). Each Transformer layer has an input embedding dimension of 512 and a feed-forward layer dimension of 2048. The hyper-parameters in Eq.~\ref{eq:total-loss} are set as: $\alpha=1.0$ and $\beta=1.0$, respectively.
% ($[0.0, 1.0]$ for $\alpha$, $0.1, 0.5, 1.0$ for $\beta$, $0.1, 0.5, 1.0$ for $\gamma$, and $[0.01, 0.10]$ for $\delta$). 
The temperature of the softmax function in Eq.~\ref{eq:shrink} ($\mu$) is $1.0$, while the threshold $\theta$ in the boundary predictor is set to $0.4$ during inference. All the above hyper-parameters are set through grid search based on the performance of the development set.
% \footnote{We also apply similar grid search to the hyper-parameters of the compared methods.}.
Training details please refer to Appendix~\ref{sec:appendix:imp}.

% We use the Adam optimizer~\cite{DBLP:journals/corr/KingmaB14} with a 0.002 learning rate and 10000 warm-up steps followed by the inverse square root scheduler to train our models. Label smoothing and dropout strategies are used, both set to $0.1$. The models are fine-tuned on 8 NVIDIA Tesla V100 GPUs up to 100 epochs with early stop. The batch size is set to 40000 frames per GPU.
% We save checkpoints every epoch and average the last 10 checkpoints for evaluation with a beam size of $5$.

We apply SacreBLEU\footnote{https://github.com/mjpost/sacreBLEU} for evaluation, where case-sensitive detokenized BLEU is reported. 

\subsection{Experiment results}
% \begin{table}[t]
% \small
% \begin{center}
% \setlength{\tabcolsep}{1.0mm}
% \begin{tabular}{l|c|c|c|cc}
% \toprule[1pt]
% \multirow{2}{*}{\bf{Model}} & \bf{Diff$\le$2} & \multirow{2}{*}{\bf{Speedup}} & \bf{Mem} & \multicolumn{2}{c}{\bf BLEU}\\
% & (\%) & & \bf{Usage} & En-De & En-Fr\\
% \midrule[0.5pt]
% No Shrink & -- & 1.00$\times$ & 1.00 & 26.0 & 36.8 \\
% Fix Shrink  & 36.7 & \textbf{1.06$\times$} & \textbf{0.74} & 25.4 & 36.0 \\
% CIF-Based & 70.3 & 1.04$\times$& \textbf{0.74} & 25.8 & 36.2 \\ 
% % $\;\;$+ASR MTL & \underline{81.9} &&& 26.3 & 36.8 \\
% CTC-Based & \underline{80.2} & 0.76$\times$& 1.77 & \underline{26.4} & \underline{36.9} \\
% % $\;\;$+ Weighted & 79.1 &&& 26.3 & 36.8 \\
% Boundary-Based & \textbf{81.9} &\textbf{1.06$\times$}& \underline{0.78} & \textbf{26.7} & \textbf{37.4} \\
% % $\;\;$+ Forced & \textbf{99.0} &&& \underline{26.6} & \underline{37.3} \\
% % \midrule[0.5pt]
% % Oracle Results & 99.7 & 26.5 \\
% \bottomrule[1pt]
% \end{tabular} 
% \end{center}
% \vskip -1.0em
% \caption{
% The results of shrinking-based methods and the corresponding shrinking quality, evaluated with length differences between the shrunk representations and transcriptions. Diff$\le$2 means the length differences are less than or equal to 2. The speedup and memory usage are both tested with a batch size of 16, and we only display the relative values for clear comparison. }
% \vskip -1.0em
% \label{tab:shrink-quality}
% \end{table}

\begin{table}[t]
\small
\begin{center}
\setlength{\tabcolsep}{1.0mm}
\begin{tabular}{l|c|c|c|cc}
\toprule[1pt]
\multirow{2}{*}{\bf{Model}} & \bf{Diff$\le$2} & \multirow{2}{*}{\bf{Speedup}} & \bf{Mem} & \multicolumn{2}{c}{\bf BLEU}\\
& (\%) & & \bf{Usage} & En-De & En-Fr\\
\midrule[0.5pt]
No Shrink & -- & 1.00$\times$ & 1.00 & 26.0 & 36.8 \\
Fix Shrink  & 36.7 & \textbf{1.06$\times$} & \textbf{0.74} & 25.4 & 36.0 \\
CIF-Based & 70.3 & \underline{1.04$\times$}& \textbf{0.74} & 25.8 & 36.2 \\ 
CTC-Based & \underline{80.2} & 0.76$\times$& 1.77 & \underline{26.4} & 36.9 \\
\midrule[0.5pt]
Boundary-Based & \multirow{3}{*}{\textbf{81.9}} &\multirow{3}{*}{\textbf{1.06$\times$}}& \multirow{3}{*}{\underline{0.78}} & \textbf{26.7} & \textbf{37.4} \\
$\;\;$- Forced Train &&&& \underline{26.4} & \underline{37.1} \\
$\;\;$$\;\;$- Blank Label &&&& 26.3 & 36.6 \\
% \midrule[0.5pt]
% Oracle Results & 99.7 & 26.5 \\
\bottomrule[1pt]
\end{tabular} 
\end{center}
\vskip -1.0em
\caption{
The results of shrinking-based methods and the corresponding shrinking quality, evaluated with length differences between the shrunk representations and transcriptions. Diff$\le$2 means the length differences are less than or equal to 2. The speedup and memory usage are both tested with a batch size of 16, and we only display the relative values for clear comparison. }
\vskip -1.0em
\label{tab:shrink-quality}
\end{table}

% \begin{table}[t]
% \small
% \begin{center}
% \setlength{\tabcolsep}{1.0mm}
% \begin{tabular}{l | c | c c}
% \toprule[1pt]
% \multirow{2}{*}{\bf{Model}} & \multirow{2}{*}{\bf{Diff$\le$2 (\%)}} & \multicolumn{2}{c}{\bf BLEU}\\
% & & En-De & En-Fr\\
% \midrule[0.5pt]
% No Shrink (Baseline) & -- & 26.0 & 36.8 \\
% \midrule[0.5pt]
% Fix-Shrink (Every 3 frames) & 36.7 & 25.4 & 36.0 \\
% CTC-Based~\cite{DBLP:journals/corr/abs-2010-14920} & 80.2 & 26.4 & 36.9 \\
% Weighted~\cite{zeng-etal-2021-realtrans} & 79.1 & 26.3 & 36.8 \\
% CIF-Based~\cite{dong2021unist} & 81.9 & 26.3 & 36.8 \\
% \midrule[0.5pt]
% Boundary-Based (AdaTranS) & 81.9 & \textbf{26.7} & \textbf{37.4} \\
% $\;\;$+ Source Text Guidance & \textbf{99.0} & 26.6 & 37.3 \\
% % \midrule[0.5pt]
% % Oracle Results & 99.7 & 26.5 \\
% \bottomrule[1pt]
% \end{tabular} 
% \end{center}
% \vskip -1.0em
% \caption{
% The results of shrinking-based methods and the corresponding shrinking quality, evaluated with length differences between the shrunk representations and transcriptions. Diff$\le$2 means the length differences are less than or equal to 2. }
% \vskip -1.0em
% \label{tab:shrink-quality}
% \end{table}
Table \ref{tab:shrink-quality} compares different shrinking-based methods in terms of quality and efficiency. 
% We do not apply any modality alignment losses (including KD) for simplicity and a fair comparison. 
Besides translation quality, we use length differences between the shrunk representations and the corresponding transcriptions to evaluate shrinking quality following \citet{zeng-etal-2021-realtrans}. 
We use inference speedup and memory usage to evaluate the efficiency.

For comparisons, the Fix-Shrink method shrinks the speech features with a fixed rate (e.g. every 3 frames). The CIF-Based method~\cite{dong-etal-2022-learning} is based on a continuous integrate-and-fire mechanism.
The CTC-Based method~\cite{DBLP:journals/corr/abs-2010-14920} shrinks features based on CTC greedy paths. 
% while \citet{zeng-etal-2021-realtrans} proposes to further use weighted average hidden states as shrunk representations (noted as ``+ Weighted'').
% which slows down the system's inference speed and introduces extra memory cost. 
As can be seen, poor shrinking (Fix-Shrink and CIF-based) hurts the performance, although with better efficiency. 
The boundary-based shrinking used in AdaTranS and the CTC-based method achieve better shrinking quality, with performance improved. 
However, CTC-Based method hurts the inference efficiency (lower inference speed and higher memory usage) as they introduce extra computation cost producing greedy CTC path in a large source vocabulary.
% ; while CIF-Based method also hurts the performance without extra multi-task learning objective (``ASR MTL'', which increses training cost).
Our method performs the best in both shrinking and translation quality with nice inference efficiency. This demonstrates the effectiveness of our method. 

On the other hand, we also notice that removing forced training trick (``-Forced Train'') or weighted-shrinking (i.e., ``-Blank Label'', simply average the frame representations rather than use Eq.~\ref{eq:shrink}) will affects the translation quality, showing the effectiveness of these two components.

% On the other hand, according to the results of ``+ Forced'' which uses the forced trick introduced in Section~\ref{sec:model:align} during inference, it seems that there is no need to further reduce the length differences\footnote{We further verify it by using a forced alignment tool to extract the alignment between the speech and transcriptions and guide the shrinking of the speech features. Still no further improvement is observed.}. This may suggest that in addition to the length differences, representations need refinement as well, which inspires the modality alignment objectives.

\paragraph{Adopting Alignment Objectives.}
\begin{table}[t]
\small
\begin{center}
\setlength{\tabcolsep}{1.7mm}
\begin{tabular}{l|ccc}
\toprule[1pt]
\multirow{2}{*}{\bf{Model}} &  \multicolumn{3}{c}{\bf BLEU} \\
& En-De & En-Fr & En-Ru  \\
\midrule[0.5pt]
MT & 34.4 & 44.9 & 21.3  \\
Cascaded Model & 28.1 & 37.0 & 17.6 \\
STPT$^{\dagger}$~\cite{tang-etal-2022-unified} & 29.2 & 39.7 & -- \\
SpeechUT$^{\dagger}$~\cite{DBLP:journals/corr/abs-2210-03730} & 30.1 & 41.4 & --\\
\midrule[0.5pt]
JT-S-MT~\cite{tang-etal-2021-improving} & 26.8 & 37.4 & --  \\
Chimera~\cite{han-etal-2021-learning} & 27.1 & 35.6 & 17.4 \\
XSTNet~\cite{DBLP:conf/interspeech/YeW021} & 27.1 & 38.0 & 18.5 \\
SATE~\cite{xu-etal-2021-stacked} & 28.1 & -- & -- \\
STEMM~\cite{fang-etal-2022-stemm} & \textbf{28.7} & 37.4 & 17.8\\
ConST~\cite{ye-etal-2022-cross} & 28.3 & 38.3 & 18.9 \\
\midrule[0.5pt]
AdaTranS & \textbf{28.7} & \textbf{38.7} & \textbf{19.0} \\
\bottomrule[1pt]
\end{tabular} 
\end{center}
\vskip -1.0em
\caption{Comparisons with the SOTA models. The first two rows are results with our pre-trained MT model and ASR model. Models marked with $^{\dagger}$ are trained with complex speech-text joint pre-training techniques.}
\vskip -1.0em
\label{tab:main-compare-short}
\end{table}

AdaTranS can be further improved with objectives that aligning speech and text representations (i.e. bridging the representation space gap introduced in Section~\ref{sec:intro}).
Table~\ref{tab:main-compare-short} shows the results of AdaTranS equipped with Cross-model Contrastive~\cite{ye-etal-2022-cross} and knowledge distillation guided by MT. The results show that AdaTranS achieves competitive results in all the three datasets compared to previous state-of-the-art models.

\paragraph{Influence of the Boundary Threshold.}
\begin{figure}[t]
\centering
\includegraphics[width=0.48\textwidth]{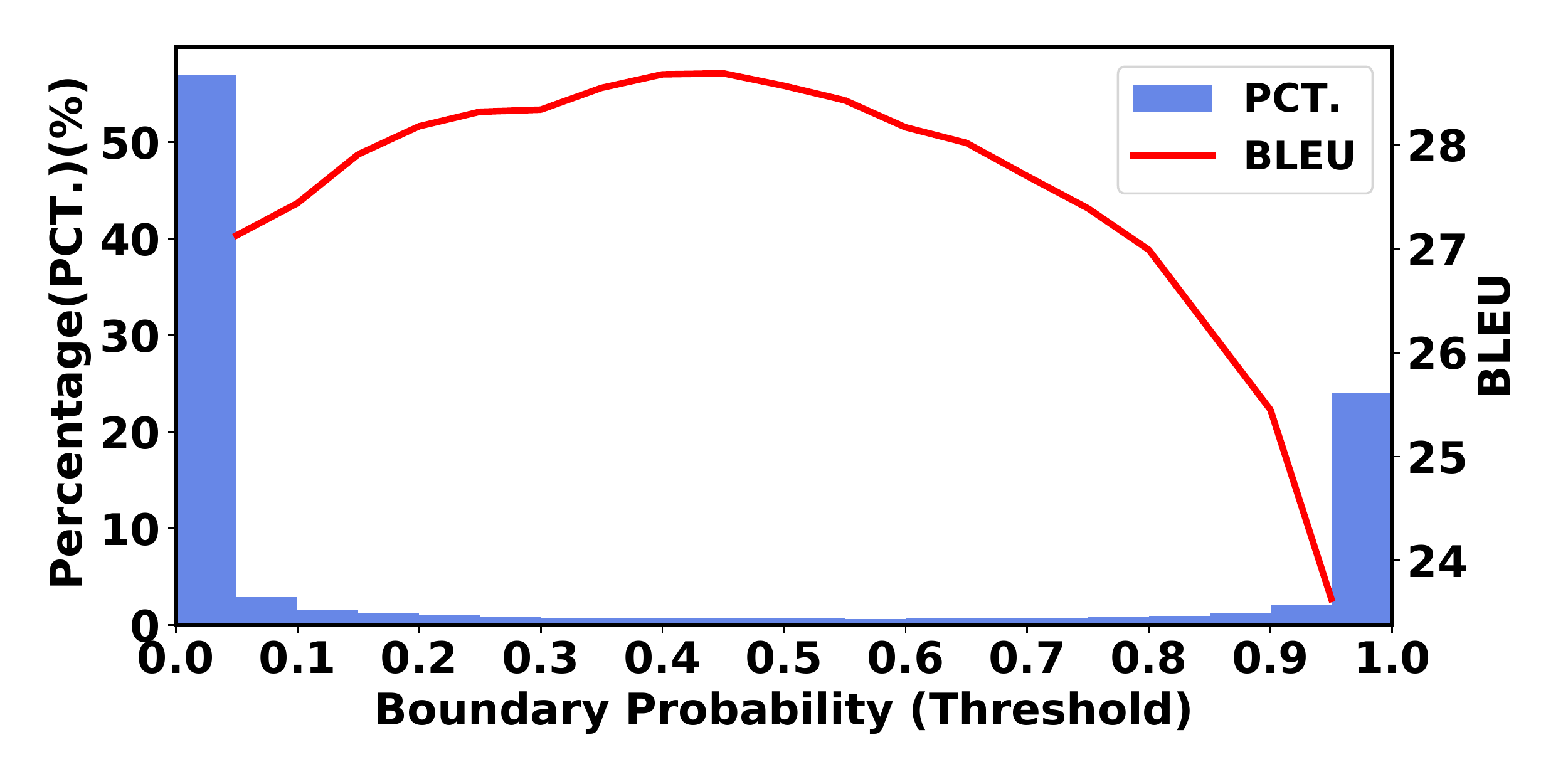}
\vskip -0.5em
\caption{
The distribution of probabilities for boundary prediction, and the corresponding model performance (BLEU) when using different values as threshold.
}
\vskip -1em
\label{fig:thres}
\end{figure}
We also examine the effects of the threshold $\theta$ for the boundary predictor. Figure~\ref{fig:thres} shows the distribution of the predicted boundary probability (i.e. $p_t(\text{<BD>})$) for each frame in the MUST-C En--De test set.
% \llyreplace{As can be seen, most of them are distinguishable, with the boundary probability smaller than $0.1$ or larger than $0.9$. Only a small portion of them have a probability between $0.2$ and $0.8$. Nevertheless, they heavily affect the final BLEU scores. Also shown is the change of BLEU scores when using different values as boundary predictor threshold. We can observe that only when the threshold is around $0.4$ can the model achieve the best performance. Different thresholds might result in a decrease in up to $5$ BLEU scores.}
We find that the boundary predictor is confident ($< 0.1$ and $> 0.9$) in most cases. However, even though only a small portion of predictions are in the range of $[0.1,0.9]$, they significantly affect the BLEU scores when the threshold changes (the red line in Figure \ref{fig:thres}). The model achieves the best performance when the threshold is around $0.4$.

% \paragraph{Influence of the Acoustic Encoder.}
% We use the Conv-Transformer (ConvT)~\cite{DBLP:conf/interspeech/HuangHYC20} as our acoustic encoder in the aforementioned experiments. 
% Here we also examine our shrinking method on different acoustic encoders.
% % We examine whether our boundary-based shrinking method still works well on acoustic encoders of different architectures. 
% We choose the S2T Transformer (S2T)~\cite{wang-etal-2020-fairseq}, a kind of vanilla transformer with 2 convolution layers at the beginning, and the Conformer (CFM)~\cite{DBLP:conf/interspeech/GulatiQCPZYHWZW20}, a more complicated convolution-augmented transformer model. We also remove the relative positional encoding (- Rel) from the Conv-Transformer as a comparison. Table~\ref{tab:acoustic-enc} shows the results of these variants. We can see that compared to the ST baseline without shrinking (ST-B) and the CTC-based shrinking model (CTC-Shr), AdaTranS always performs the best on different acoustic encoders, demonstrating the generalizability of our method. We also have two further observations. First, the performance of the pre-trained ASR seems not consistent with the ST performance. The second is that more complicated acoustic encoders (i.e. more parameters) are beneficial.
% \input{tables/acoustic_enc}

\section{Conclusion}
This work proposes a new end-to-end ST model called AdaTranS, which uses a boundary predictor trained by signals from CTC output probabilities, to adapt and bridge the length gap between speech and text. 
Experiments show that AdaTranS performs better than other shrinking-based methods, in terms of both quality and efficiency.
It can also be further enhanced by modality alignment objectives to achieve state-of-the-art results.
% \llydelete{modality alignment with a forced training trick and a contrastive training objective, so that the ST model can better inherit knowledge from the pre-trained models}
% to improve the knowledge in pre-trained models transferring from the text domain to the speech domain. Experiments show that AdaTranS performs better than other related methods.
% \llydelete{, and has speed and memory advantage during inference. Further analyses also show that our method indeed bridges the modality gap to some extent and therefore yields better results.}
% In addition, it has a faster inference speed and lower memory usage than the baselines. Further analyses show its generalizability to different architectures.

\bibliography{acl}
\bibliographystyle{acl_natbib}

\newpage
\appendix

\section{Data Statistics.}\label{sec:appendix:data}
Table~\ref{tab:stat} shows the data statistics of the used datasets. ST datasets are all from MUST-C, and LibriSpeech serves as extra ASR data. MT data either comes from OpenSubtitles2018 or WMT14 following previous work settings.
\begin{table}[ht]
\small
\begin{center}
\setlength{\tabcolsep}{1.4mm}
\begin{tabular}{l|c c c}
\toprule[1pt]
Corpus&ST(Hours/\#Sents)&ASR(Hours)&MT(\#Sents)\\
\midrule[0.5pt]
En--De&408/234K&960&18M(OS)\\
En--Fr&492/280K&960&18M(WMT)\\
En--Ru&489/270K&960&2.5M(WMT) \\
\bottomrule[1pt]
\end{tabular} 
\end{center}
\vskip -1.0em
\caption{
The statistics for the three language pairs. OS: OpenSubtiles2018. WMT: WMT14.
}
\vskip -1.0em
\label{tab:stat}
\end{table}

\section{Implementation Details.}\label{sec:appendix:imp}
\paragraph{Data Preprocessing.}
We use 80D log-mel filterbanks as speech input features, which are calculated with 25ms window size and 10ms step size and normalized by utterance-level Cepstral Mean and Variance Normalization (CMVN). All the texts in ST and MT data are preprocessed in the same way, which are case-sensitive with punctuation preserved. We filter out samples with more than 3000 frames, over 256 tokens, or whose ratios of source and target text lengths are outside the range [2/3, 3/2]. We use SentencePiece\footnote{https://github.com/google/sentencepiece}~\cite{kudo-richardson-2018-sentencepiece} to generate subword vocabularies for each language pair. Each vocabulary is learned on all the texts from ST and MT data and shared across source and target languages, with a total size of 16000.

\paragraph{Training Details.}
We train all the models using Adam optimizer~\cite{DBLP:journals/corr/KingmaB14} with a 0.002 learning rate and 10000 warm-up steps followed by the inverse square root scheduler. Label smoothing and dropout strategies are used, both set to 0.1. The models are fine-tuned on 8 NVIDIA Tesla V100 GPUs with 40000 steps. The batch size is set to 40000 frames per GPU.
We save checkpoints every epoch and average the last 10 checkpoints for evaluation with a beam size of 10.

\section{More Analysis.}\label{sec:appendix:ana}
\paragraph{Better Source-Target Alignment.}
\begin{figure}[ht]
\centering
\includegraphics[width=0.48\textwidth]{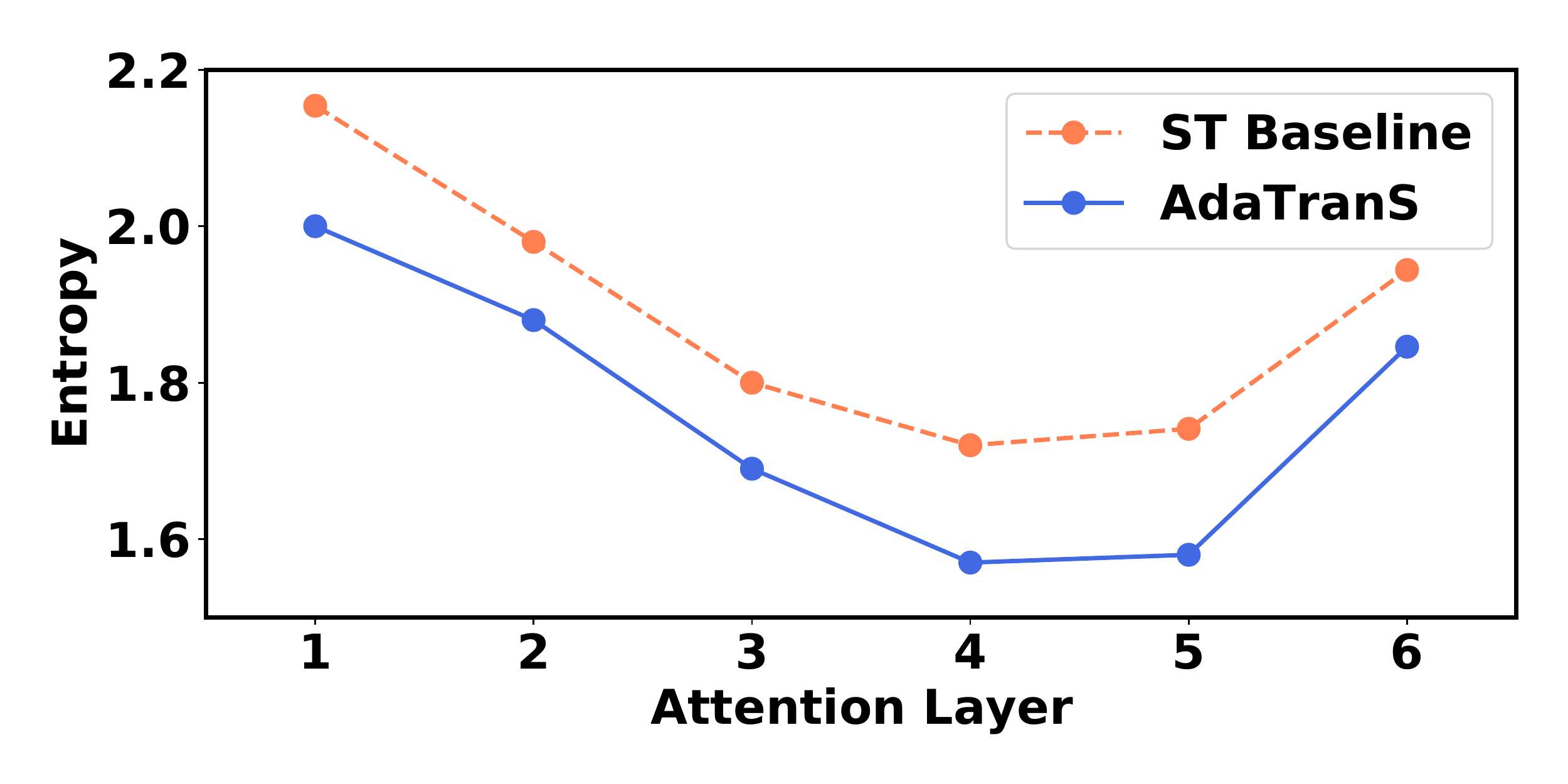}
\vskip -0.5em
\caption{
The attention entropy of each attention layer for end-to-end ST baseline and our model.
}
\vskip -1em
\label{fig:entropy}
\end{figure}
We evaluate the entropy of the cross attention from the ST baseline and AdaTranS\footnote{To fairly compare, we also shrink the speech features of the ST baseline with the same boundaries detected by our boundary predictor.}. Let $\alpha_{ij}$ be the attention weight for a target token $y_i$ and a source speech feature (after shrinking) $x_j$, the entropy for each target token is defined as $E_i = -\sum\nolimits_{j=1}^{|x|} \alpha_{ij} \log \alpha_{ij}$. We then average the attention entropy of all target tokens in the test set. Lower entropy means the attention mechanism is more confident and concentrates on the source-target alignment. Figure~\ref{fig:entropy} shows the entropy of different decoder layers. AdaTranS exhibits consistently lower entropy than the ST baseline. This means that our shrinking mechanism improves the learning of attention distributions.

\paragraph{Influence of Text Input Representations.}
\begin{table}[ht]
\small
\begin{center}
\setlength{\tabcolsep}{0.8mm}
\begin{tabular}{l|c c c}
\toprule[1pt]
& \bf{SPM}  & \bf{SPM w/o Punct} & \bf{Phoneme} \\
\midrule[0.5pt]
\underline{\bf{MT}} &&& \\
1. Only MUST-C Data & 30.7 & 28.3 & 28.2 \\
2. PT with Extra Data & 34.4 & 31.6 & 29.5 \\
\midrule[0.5pt]
\underline{\bf{ST}} &&& \\
Initialized with Model 2 & 26.0 & 25.8 & 25.6 \\
\bottomrule[1pt]
\end{tabular} 
\end{center}
\vskip -1.0em
\caption{
BLEU scores of different text input representations in MUST-C En-De. SPM means using the subword units learned from a sentencepiece model, while ``w/o Punct'' indicates that punctuation is removed.
% ``SPM w/o Punct'' indicates that we further remove the punctuation in the text input.
}
\vskip -1.0em
\label{tab:text_input}
\end{table}
Representing text input with phonemes helps reduce the differences between speech and text~\cite{tang-etal-2021-improving,DBLP:conf/icassp/TangPWMG21}. However, word representations and punctuation are important for learning semantic information, which are usually ignored when phonemes are used in prior works. Table~\ref{tab:text_input} shows the MT results when using different text input representations, together with the ST performance that is initialized from the corresponding MT model.
% \llydelete{``SPM'' means using the subword units learned from a sentencepiece model, while ``SPM w/o Punct'' indicates that we further remove the punctuation in the text input.}
% The results show that text input representations are crucial to the MT performance, especially when large amount of data is used for training. \llyreplace{And this performance difference would also affect the ST performance that initialized with the models. Therefore, we choose to use subword units with punctuation preserved in our experiments.}
We can observe that the performance of downstream ST model is affected by the pre-trained MT model. Therefore, instead of following prior phoneme-level work for pre-training the MT, in this work we use subword units with punctuation and incorporate the shrinking mechanism to mitigate the length gap.

\end{document}